\documentclass[11pt]{article}
\usepackage{coling2016}
\usepackage{times}
\usepackage{url}
\usepackage{latexsym}
\usepackage{amssymb}
\usepackage{tikz}
\usetikzlibrary{decorations.pathreplacing}
\usepackage{fancyvrb}
\usepackage{graphicx}
\usepackage{xcolor}
\usepackage{url}
\usepackage{multicol}
\usepackage{footnote}
\makesavenoteenv{tabular}
\makesavenoteenv{table}
\usepackage{listings}% http://ctan.org/pkg/listings
\definecolor{pinegreen}{rgb}{0.0, 0.47, 0.44}
\title{Exploiting Sentence and Context Representations in Deep Neural Models for Spoken Language Understanding}

\author{Lina M.~Rojas-Barahona, Milica~Ga{\v s}i{\' c},
Nikola~Mrk{\v s}i{\' c}, Pei-Hao~Su \\ {\bf Stefan~Ultes, Tsung-Hsien~Wen and Steve~Young}\\
Department of Engineering, University of Cambridge, Cambridge, UK\\
{lmr46, mg436, nm480, phs26, su259, thw28, sjy}@cam.ac.uk}

\date{}

\begin{document}
\maketitle
\begin{abstract}
This paper presents a deep learning architecture for the semantic decoder component of a Statistical Spoken Dialogue System. In a slot-filling dialogue, the semantic decoder predicts the dialogue act and a set of slot-value pairs from a set of n-best hypotheses returned by the Automatic Speech Recognition. 
Most current  models for spoken language understanding assume (i) word-aligned semantic annotations as in sequence taggers and (ii) delexicalisation, or a mapping of input words to domain-specific concepts using heuristics that  try to capture morphological variation but that do not scale to other domains nor to language variation (e.g., morphology, synonyms, paraphrasing ). In this work the semantic decoder is trained using unaligned semantic annotations and it uses distributed semantic representation learning to overcome the limitations of explicit delexicalisation.  The proposed  architecture uses a convolutional neural network for the  \textit{sentence representation} and a long-short term memory network for the \textit{context representation}. Results are presented for the publicly available DSTC2 corpus and an In-car corpus which is similar to DSTC2 but has a significantly higher word error rate (WER).
\end{abstract}

%\section{Credits}

\section{Introduction}
\label{intro}

\blfootnote{
    %
    % for review submission
    %
    %
    % % final paper: en-uk version (to license, a licence)
    %
    \hspace{-0.65cm}  % space normally used by the marker
    This work is licensed under a Creative Commons 
    Attribution 4.0 International Licence.
    Licence details:
    \url{http://creativecommons.org/licenses/by/4.0/}
    % 
    % % final paper: en-us version (to licence, a license)
    %
    % \hspace{-0.65cm}  % space normally used by the marker
    % This work is licenced under a Creative Commons 
    % Attribution 4.0 International License.
    % License details:
    % \url{http://creativecommons.org/licenses/by/4.0/}
}
%Deep Learning has been used for
In most existing work on Spoken Language Understanding (SLU),  semantic decoding is usually seen as a sequence tagging problem with models trained and tested on datasets with word-level annotations \cite{tur2013semantic,mesnil2015using,yao2013recurrent,5947649,deoras2013deep,sarikaya2014application}.  Spoken language understanding from \textit{unaligned data}, in which utterances are annotated with an abstract semantics, faces the additional challenge of not knowing which specific words are relevant for extracting the semantics. This problem was tackled in~\cite{zhou2011learning}, by using conditional random fields (CRFs) driven by finely-tuned hand-crafted features.
Other discriminative approaches that deal with unaligned data use some form of \textit{delexicalisation} or mapping of the input to known ontological concepts \cite{Henderson2012a,Henderson2014b}. The main disadvantage of delexicalisation is the difficulty in  scaling it, not only to larger and more complex dialogue domains but also to handle the many forms of language variation.

We propose in this paper a semantic decoder that learns from unaligned data (Figure~\ref{f:dial}) and that exploits rich semantic distributed word representations instead of delexicalisation. The semantic decoder predicts the dialogue act and the set of slot-value pairs from a set of n-best hypotheses returned by an automatic speech recognition (ASR).  The prediction is made in two steps. First, a deep learning architecture is used for the joint prediction of dialogue acts and the presence or absence of slots. Second, the same architecture is reused for predicting the values of the slots that were detected by the first joint-classifier. The deep architecture combines  sentence and context representations. A convolutional neural network (CNN)~\cite{collobert2011natural} is used to generate the sentence representation, while a long-short term memory (LSTM) network~\cite{hochreiter1997long} is used to generate the context representation.  A non-linear function then combines the top layers of these neural networks and distinct softmax layers are used to predict the dialogue act and slots in the first joint model. In the second model, a single softmax predicts the possible values for each slot.

\begin{figure}[ht]

%\begin{Verbatim}[fontsize=\scriptsize,commandchars=\\\{\}]
\noindent
\begin{tabular}{ll}
\scriptsize{\textbf{voip-922209b777-20130325\_155209}}&\\
\scriptsize{\textbf{System: }Hello , welcome to the Cambridge restaurant system? You can ask for restaurants by area,} \\[-1ex]
    \scriptsize{price range or food type. How may I help you?}&
    \scriptsize{$\rightarrow$\textbf{\textit{welcomemsg}}}\\
\colorbox{blue!20}{\scriptsize{\textbf{User: }  i am looking for a moderately priced restaurant in the north part}}&
    \scriptsize{$\rightarrow$ \color{pinegreen}\textbf{\textit{inform(area=north,pricerange=moderate)}}\color{black}}\\
\scriptsize{\textbf{System: }meghna is a nice restaurant in the north of town in the moderate price range} &
    \scriptsize{$\rightarrow$ \textbf{\textit{offer(name=meghna),inform(pricerange=moderate,}}} \\[-1ex] &  \hspace{0.3cm}\scriptsize{\textbf{\textit{area=north)}}}\\
\colorbox{blue!20}{\scriptsize{\textbf{User: } is everything else}} &  \scriptsize{$\rightarrow$ \color{pinegreen}\textbf{\textit{reqalts()}}}\color{black}\\
\scriptsize{\textbf{System: }golden wok is a nice restaurant in the north of town in the moderate price range} &
   \scriptsize{$\rightarrow$ \textbf{\textit{offer(name=golden...),inform(pricerange=moderate,}}} \\[-1ex] &  \hspace{0.3cm}\scriptsize{\textbf{\textit{area=north)}}}\\
\colorbox{blue!20}{\scriptsize{\textbf{User: } can i get the phone number}} &  \scriptsize{$\rightarrow$ \color{pinegreen}\textbf{\textit{request(slot=phone)}}\color{black}}\\
\scriptsize{\textbf{System: }The phone number of golden wok is ...} &
  \scriptsize{$\rightarrow$  \textbf{\textit{offer(name=golden wok),inform(phone=01..)}}} \\
\colorbox{blue!20}{\scriptsize{ \textbf{User: } \textit{type of food do they serve}}} &  \scriptsize{$\rightarrow$ \color{pinegreen}\textbf{\textit{request(slot=food)}}\color{black}}

\end{tabular}
%\end{Verbatim}

\caption{Excerpt from a dialogue in the DSTC2 corpus. The top-best ASR hypothesis is shown highlighted on the left, and the corresponding user semantics is shown highlighted on the right.}
\label{f:dial}
\end{figure}

Our models are evaluated on two datasets DSTC2~\cite{henderson2014second} and In-car~\cite{tsiakoulis2012statistical} using accuracy, f-measure and the Item Cross Entropy (ICE) score~\cite{thomson2008evaluating}. We show that these models outperform previous proposed models, without using manually designed features and without any preprocessing of the input (e.g., stop words filtering, delexicalisation). They do this
by exploiting distributed word representations and
we claim that this allows semantic decoders to be built that can easily scale to larger and more complex dialogue domains.

The remainder of this paper is structured as follows. We first present related work in Section \ref{s:relwork} and then we describe our architecture in Section~\ref{s:dlsemi}. We describe the experimental setup in~\ref{s:exp} and the evaluation results are introduced in Section~\ref{s:eval}. Finally, we present conclusions and future work in Section~\ref{s:disc}.
 
\section{Related Work}
\label{s:relwork}
Sequence tagging discriminative models such as CRFs and sequence neural networks have been widely explored for spoken language understanding. 
%In these models Spoken Language Understanding is seen as a sequence tagging.  
For instance, Recurrent Neural Networks have been proposed in \cite{yao2013recurrent,mesnil2015using} and
generative  Deep Neural Networks consisting of  a composition of Restricted Boltzmann Machines (RBM) have been studied by~\cite{5947649,deoras2013deep,sarikaya2014application}. 
A combination of neural networks and triangular CRFs is presented in~\cite{celikyilmaz2010convolutional}, in which
a convolutional neural network is used for extracting the input features of a triangular CRF in order to perform  joint intent detection and slot filling.
All these models use word-level semantic annotations. However, providing these word-level semantic annotations is costly since it requires specialised annotators.
\cite{zhou2011learning} has proposed learning CRFs from unaligned data, however they use manually tuned lexical or syntactic features. In this work we avoid the need for word-level annotation by exploiting distributed word embeddings and using deep learning for feature representation. 

Convolutional Neural Networks (CNNs) have been used previously for sentiment analysis~\cite{kim2014convolutional,kalchbrenner2014convolutional} and in this work we explore a similar CNN to the one presented by~\newcite{kim2014convolutional} for generating a sentence representation. However unlike~\newcite{kim2014convolutional}, the input in not a single well formed sentence but a set of ill-formed ASR hypotheses. Additionally, the softmax layer used for binary classification (i.e., positive or negative sentiment) is replaced by a softmax layer for multiclass dialogue act prediction and a further softmax layer is added for each distinct slot in the domain. \cite{chen2015learning} proposed a CNN for generating intent embeddings in SLU, which uses tri-letter input vectors. Instead, in this paper the models are initialised with GloVe word embeddings~\cite{pennington2014glove}. These GloVe embeddings were trained in an unsupervised fashion on a large amount of data to model the contextual  similarity and correlation between words. Chen and He's model aims to learn the embeddings for utterances and intents such that utterances with similar intents are close to each other in the continuous space. Although we share the same spirit, we use sentence embeddings not only for intent (or dialogue act) recognition but also for slot-filling within a dialogue system and we combine them with embeddings for dialogue context.

Approaches for adaptive SLU have been proposed in~\cite{ferreira2015online,zhu2014semantic}, however they focused more on domain adaptation on top of an existing SLU component. Moreover, they use classical discriminative models for SLU such as CRFs and SVMs that require manually designed features. In contrast, the focus of this paper is to exploit deep learning models for SLU, which learn  feature representations automatically.

Recently, some researchers have focused on mapping word level hypotheses directly to beliefs without using an explicit
semantic decoder step~\cite{Henderson2014b,Mrksic:15}. These systems track the user's goal through the course of the dialogue by maintaining a distribution over slot-value pairs. Such systems are interesting, but it is not clear that they can be scaled to very large domains due to the constraint of delexicalisation.  Furthermore, they still require an explicit semantic decoding layer for domain identification and general 
 \textit{Topic Management}. %Distributed vector representations on the contrary can encode similarities across domains.

\section{Deep Learning Semantic Decoder}
\label{s:dlsemi}
\begin{figure}[ht]
\centering
\includegraphics[scale=0.4]{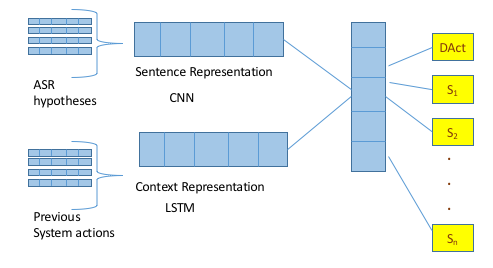}
\caption{Combination of sentence and context representations for the joint prediction of dialogue acts and slots.}
\label{f:dlarchi}
\end{figure}

We split the task of  semantic decoding into two steps: (i) training a joint model for predicting the dialogue act and presence or absence of slots and (ii) predicting the values for the most probable slots detected in (i). 
%The benefits of joint-inference have been shown for different NLP tasks such as entity-relation classification, correlation and parsing~\cite{Singh2013}. The main advantage of joint-inference is that parameters are shared between predictors, thus weights can be adjusted based on their mutual influence. For instance, dialogue acts might influence slots prediction, since some slots are likely to appear under the scope of a specific dialogue act (e.g., the slot \textit{food} is more likely to appear under \textit{Inform} rather than \textit{Request}).
As shown in Figure~\ref{f:dlarchi}, we use the same deep learning architecture in both steps for combining sentence and context representations to generate the final hidden unit that feeds one or many softmax layers.  In the first step, as shown in the Figure, there are distinct softmax layers for the joint optimisation of the dialogue act and each possible slot.  In the second step there is a single softmax layer that predicts the value of each specific slot. In the following we explain this architecture in more detail.

%\subsection{Deep Architecture}
%The proposed architecture combines sentence and context representations as follows.
\subsection{Sentence Representation}
\label{ss:srepr}
A CNN is used for generating the hypothesis representation, then these representations are weighted by their confidence scores and then summed up to obtain the sentence representation (Figure~\ref{f:cnn}). 

The CNN is a variant of {~\cite{kim2014convolutional}, in which the inputs are the word vectors in each \textit{ASR hypothesis}. Let $x_i$ be a $k-$dimensional word embedding for the $i$-th word in a hypothesis.
A hypothesis of length $m$ is represented as:
	$x_{1:m}=\mathbf{x}_1 \bigoplus \mathbf{x}_2  \bigoplus ... \bigoplus \mathbf{x}_m$
where $\bigoplus$ is the concatenation operator. A convolutional operation is applied to a window of $l$ words to produce a new feature.
\begin{equation}
 c_i=f(\mathbf{w}\cdot\mathbf{x}_{i:i+l-1}+b)
 \label{eq:conv}
\end{equation} 
where $f$ is the hyperbolic tangent function; $w \in \mathbb{R}^{lk}$ is a filter  applied to a window of $l$ words and $b \in \mathbb{R}$ is a bias term.
The filter is applied to every window of words in the sentence to produce a feature map.
\begin{equation}
	\mathbf{c}=[ c_1, c_2, ... , c_{n-l+1}]
	\label{eq:featmap}
\end{equation}
with  $\mathbf{c} \in \mathbb{R}^{n-l+1}$. A max pooling operation is then applied to give the maximum value $c=max\{\mathbf{c}\}$ as the representative feature for that filter. Multiple filters can be applied by varying the window size to obtain several adjacent features for a given hypothesis.
These features $\hat{f}_j$ for the hypothesis $j \in H$ are then multiplied by the ASR confidence score $p_j$\footnote{The posterior probability of hypothesis $j$ in the N-best list.} 
and summed over all ASR hypotheses to generate a representation for the sentence $s_t$ (Equation~\ref{eq:wsum}), as shown in Figure~\ref{f:cnn}.
\begin{equation}
 s_t= \sum_{j \in H} \hat{f}_j * p_j
 \label{eq:wsum}
\end{equation}

\begin{figure}[h]
\begin{tikzpicture}[scale=.5]
%i'm looking for uh a moderately priced restaurant
\node[text width=0.2cm] at (-1.3,8.5) {\scriptsize{i}};
\node[text width=0.2cm] at (-1.3,7.5) {\scriptsize{'m}};
\node[text width=0.2cm] at (-1.7,6.5) {\scriptsize{looking}};
\node[text width=0.2cm] at (-1.3,5.5) {\scriptsize{for}};
\node[text width=0.2cm] at (-1.3,4.5) {\scriptsize{uh}};
\node[text width=0.2cm] at (-1.3,3.5) {\scriptsize{a}};
\node[text width=0.2cm] at (-2.1,2.5) {\scriptsize{moderately}};
\node[text width=0.2cm] at (-1.3,1.5) {\scriptsize{priced}};
\node[text width=0.2cm] at (-1.9,0.5) {\scriptsize{restaurant}};
\draw[thick, scale=1] (0, 0) grid (6, 9);
\draw[line width=.8pt,very thick,blue](0,0)--(0,2);
\draw[line width=.8pt,very thick,blue](0,2)--(6,2);
\draw[line width=.8pt,very thick,blue](6,2)--(6,0);
\draw[line width=.8pt,very thick,blue](0,0)--(6,0);

\draw[line width=.8pt,very thick,green](0,3)--(0,6);
\draw[line width=.8pt,very thick,green](0,6)--(6,6);
\draw[line width=.8pt,very thick,green](6,6)--(6,3);
\draw[line width=.8pt,very thick,green](0,3)--(6,3);

\draw[line width=.8pt,very thick,blue](0,7)--(0,9);
\draw[line width=.8pt,very thick,blue](0,9)--(6,9);
\draw[line width=.8pt,very thick,blue](6,9)--(6,7);
\draw[line width=.8pt,very thick,blue](0,7)--(6,7);

\draw[line width=.8pt] (1, 10)--(7, 10);
%\draw[line width=.8pt] (1, 0)--(1, 10);
%\draw[line width=.8pt] (6, 1)--(7, 1);
\draw[thin, scale=1] (7, 1) grid (7, 10);
\draw[line width=.8pt, black](7,10)--(7,10);
\draw[line width=.8pt,black](7,1)--(7,10);
\draw[thin, scale=1] (2, 11) grid (8, 11);
\draw[thin, scale=1] (8, 2) grid (8, 11);
\draw[line width=.8pt,thick,black](2,11)--(8,11);

\draw [ thick,  decoration={  brace,   mirror,  raise=0.5cm  }, decorate ](-1,0) -- (9,0) node [pos=0.5,anchor=north,yshift=-0.55cm] {\scriptsize{ASR hypotheses}};

\draw[thin] (13, 0) grid (14,7);
\draw[thin] (12, 1) grid (13, 8);
\draw[thin] (11, 2) grid (12, 9);
\draw[line width=.8pt,very thick,blue](13,0)--(14,0);\draw[line width=.8pt,very thick,blue](14,0)--(14,1);
\draw[line width=.8pt,very thick,blue](14,1)--(13,1);\draw[line width=.8pt,very thick,blue](13,0)--(13,1);

\draw[line width=.8pt,very thick,green](12,3)--(13,3);\draw[line width=.8pt,very thick,green](13,3)--(13,4);
\draw[line width=.8pt,very thick,green](12,4)--(13,4);\draw[line width=.8pt,very thick,green](12,3)--(12,4);

\draw[line width=.8pt,very thick,blue](13,6)--(14,6);\draw[line width=.8pt,very thick,blue](14,6)--(14,7);
\draw[line width=.8pt,very thick,blue](14,7)--(13,7);\draw[line width=.8pt,very thick,blue](13,6)--(13,7);

%\draw[line width=.8pt,very thick,yellow](11,4)--(12,4);\draw[line width=.8pt,very thick,yellow](12,4)--(12,5);
%\draw[line width=.8pt,very thick,orange](12,5)--(11,5);\draw[line width=.8pt,very thick,orange](11,4)--(11,5);

\draw [dashed,green] (6,3) -- (12,3);
\draw [dashed,green] (6,6) -- (13,4);
\draw [dashed,blue] (6,0) -- (13,0);
\draw [dashed,blue] (6,2) -- (13,1);

\draw [dashed,blue] (6,7) -- (13,6);
\draw [dashed,blue] (6,9) -- (13,7);
\draw [ thick,  decoration={  brace,   mirror,  raise=0.5cm  }, decorate ](11,0) -- (14,0) node [pos=0.5,anchor=north,yshift=-0.55cm] {\scriptsize{Convolutional layers}};

\draw [decorate,decoration={brace,amplitude=10pt},xshift=-4pt,yshift=0pt,blue]   (-1,9) -- (2,11) node [black,midway,xshift=-0.5cm,yshift=0.5cm] {\scriptsize {N\_best}};

%\draw[thin,rotate=20] (17, -5) grid (18, -1);
\begin{scope}[
            yshift=-53,every node/.append style={
            yslant=0.4,xslant=-0.5},yslant=0.2,xslant=-1
            ]
        % opacity to prevent graphical interference
        %\fill[white,fill opacity=0.9] (0,0) rectangle (5,5);
        \draw[ black] (19,0) grid (20,4); %defining grids
        \draw[ black] (21,1) grid (22,5); %defining grids
        \draw[ black] (23,2) grid (24,6); 
        %\draw[black,very thick] (0,0) rectangle (5,5);%marking borders
        %\fill[blue] (0.05,0.05) rectangle (0.35,0.35);
        %Idem as above, for the n-th grid:
    \end{scope}
%\draw [dashed,green] (13,3) -- (19.1 ,3.9);
%\draw [dashed,green] (13,4) -- (18.4 ,4.8);
\draw [dashed,green] (13,3) -- (20.1 ,5.1);
\draw [dashed,green] (13,4) -- (19.4 ,6);
\draw [dashed,blue] (14,7) -- (20.1 ,5.1);
\draw [dashed,blue] (14,0) -- (21,4.3);

%\draw[thin] (24, 3) grid (25,7);
\fill[red!40,ultra thick] (24,3) rectangle (25,7);%cell element
\draw [ thick,  decoration={  brace,   mirror,  raise=0.5cm  }, decorate ](23,0) -- (26,0) node [pos=0.5,anchor=north,yshift=-0.55cm] {\scriptsize{Sentence Representation:}};
\node at (25,-2) {\scriptsize{weighted sum of hyps}};

\draw [ thick,  decoration={  brace,   mirror,  raise=0.5cm  }, decorate ](16,0) -- (21,0) node [pos=0.5,anchor=north,yshift=-0.55cm] {\scriptsize{hypotheses representations}};
\draw [dashed,red!50] (18,7.8) -- (24 ,7);
\draw [dashed,red!50] (20,2.1) -- (24 ,3);
%\draw [decorate,decoration={brace,amplitude=10pt},xshift=-4pt,yshift=0pt,blue]   (22,7) -- (22,2) node [black,midway,xshift=-0.5cm,yshift=0.5cm] {};
%$\sum_{i\in hyp} h_i p_i$

\end{tikzpicture}
\caption{Sentence Representation: after applying convolution operations on the N-best list of ASR hypotheses,  the resulting hidden layers are weighted by the ASR confidence scores and summed.}
\label{f:cnn}
\end{figure}
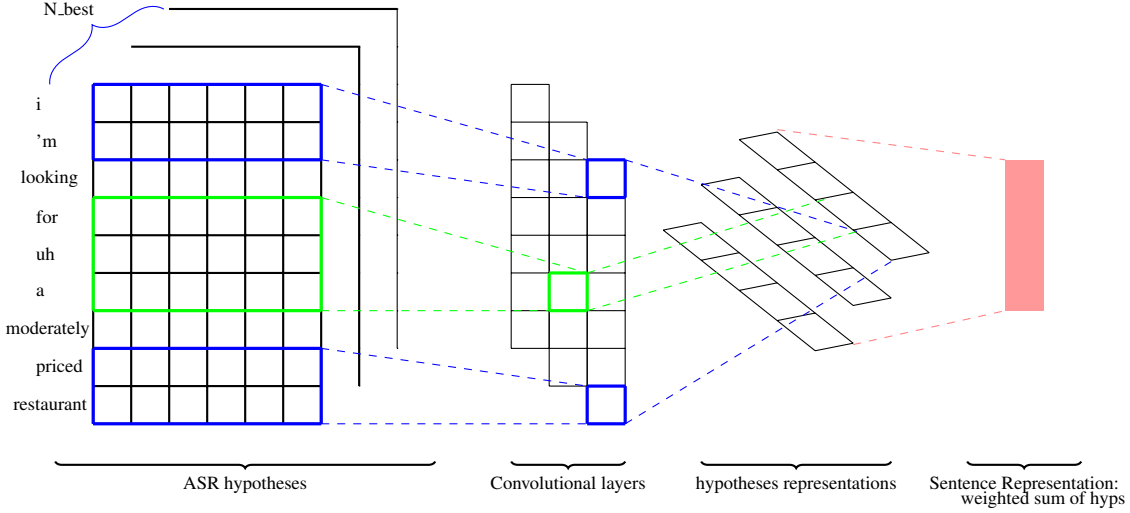

\subsection{Context Representation}
\label{ss:ctxtrepr}
An LSTM~\cite{hochreiter1997long} is used for tracking the context implied by previous dialogue system actions. The top layer of this LSTM network then provides the context representation for decoding the current input utterance. 

An LSTM is a sequence model that utilises a memory cell capable of preserving states over long periods of time.   This cell is recurrently connected to itself and it has three multiplication units, an input gate, a forget gate  and an output gate. These gating vectors are in [0,1]. The cell makes selective decisions about what information is preserved, and when to allow access to units, via gates that open and close.
The LSTM transition equations are as follows:
\begin{equation}
\begin{array}{c}
	\mathbf{i}_t=\sigma(\mathbf{W}^{(i)}\cdot x_t+ \mathbf{U}^{(i)}\cdot h_{t-1}+b^{(i)} ),\\
	\mathbf{f}_t=\sigma(\mathbf{W}^{(f)}\cdot x_t+ \mathbf{U}^{(f)}\cdot h_{t-1}+b^{(f)}),\\
	\mathbf{o}_t=\sigma(\mathbf{W}^{(o)}\cdot x_t+ \mathbf{U}^{(o)}\cdot h_{t-1}+b^{(o)}),\\
	\mathbf{u}_t=tanh(\mathbf{W}^{(u)}\cdot x_t+ \mathbf{U}^{(u)}\cdot h_{t-1}+b^{(u)}),\\
	\mathbf{c}_t=\mathbf{i}_t \bigodot \mathbf{u}_t+ \mathbf{f}_t \bigodot \mathbf{c}_{t-1},\\
	\mathbf{h}_t=\mathbf{o}_t\bigodot  tanh(\mathbf{c}_t)
\end{array}
\end{equation}
where  $h_t$ is the hidden unit at time step $t$, $x_t$ is the input at the current time step, $b$ is a bias, $\sigma$ is the logistic sigmoid function and  $\bigodot$ denotes elementwise multiplication.

As shown in Figure~\ref{f:dial}, system actions are encoded in the form of a system dialogue act plus one or more slot-value pairs.  To track the history of system actions, 
%We treat the previous system actions in the dialogue sequentially (see Figure~\ref{f:dial}). W
slots and values are treated as words and the input $x_t$ is formed from its corresponding word vectors. The length of the context can vary.  We consider all the system actions previous to the current user utterance, or a window $l$ of the previous system actions. For instance, if we are currently processing the last user input in Figure~\ref{f:dial},  in which L is the total number of system actions, we can consider all previous system actions (L=4), or the last $l$ system actions, where $l<L$.

\subsection{Combining Sentence and Context}
\label{ss:comb}
We study in this paper two ways of combining the sentence $\mathbf{s}_t$ and the context $\mathbf{h}_t$ representations. The first straightforward way is to apply a non linear function to their weighted sum:

\begin{equation}
   \mathbf{\hat{h}_t}=tanh(\mathbf{Ws}\cdot\mathbf{s}_t+\mathbf{Wc}\cdot\mathbf{h}_t)
   \label{eq:comb}
\end{equation}
The second way is to let the sentence representation be the last input to the LSTM network, then $\mathbf{\hat{h}_t} = \mathbf{h}_t$.

For classification a softmax layer is used for each prediction:
\begin{equation}
P(Y=k|\hat{h},W,b) = \frac{e^{(W_{k}\hat{h}+b_k)}}{\sum_{k\prime} e^{(W_{k\prime}\hat{h}+b_k\prime)}}
 \label{eq:softmax}
\end{equation}
where $k$ is the index of the output neuron representing one class. For dialogue act classification $k$ is one of the possible values: inform, request, offer, ... etc. For the slot prediction $k$ is either $0$ for absent or $1$ for present. For  slot-value prediction $k$ will correspond to one of the possible values for each slot. For instance, for the slot price-range the possible values are cheap, moderate, expensive and dontcare.
The result of the prediction is the most probable class:
\begin{equation}
 \hat{y}=\mbox{argmax}_k(P(Y=k|\hat{h},W,b))
\end{equation}

The back-propagation optimisation is done by minimising the negative log-likelihood loss function through stochastic gradient descent.

\section{Experimental Evaluation}
\label{s:exp}
In this section we introduce the corpora, and describe the experiments performed and the evaluation metrics used.
\subsection{Corpora}
Experimental evaluation used two similar datasets: DSTC2~\cite{henderson2014second} and In-car~\cite{tsiakoulis2012statistical}. Both corpora were collected using a spoken dialogue system which provides restaurant information system for the city of Cambridge. Users can specify restaurant suggestions by area, price-range and food type and can then query the system for additional restaurant specific information such as phone number, post code and address. The first dialogue corpus was released for the dialogue state tracking challenge and we use here the semantic annotations that were also provided~\footnote{The DSTC2 corpus is publicly available in: \url{http://camdial.org/~mh521/dstc/}}. The trainset has $2118$  dialogues and $15611$ turns in total while the testset has $1117$ dialogues  and $9890$ turns in total. 

The second corpus contains dialogues collected under various noisy in-car conditions. In a stationary car with the air conditioning fan on and off, in a moving car and in a car simulator~\cite{tsiakoulis2012statistical}~\footnote{This corpus has been obtained in an industry funded project and therefore it is not available for public use.}. 
The trainset has $1508$ dialogues and $10532$ turns in total and the testset has $641$ dialogues and $4861$ turns in total.  Because of the noise, the average word error rate (WER = $37\%$) is significantly higher than for DSTC2 (around $29\%$).

\subsection{Hyperparameters and Training}
Dropout was used on the penultimate layers  of both the CNN and the LSTM networks to prevent co-adaptation of hidden units by randomly dropping out a proportion of the hidden units during forward propagation~\cite{hinton2012improving}. The models were implemented in Theano~\cite{bastien2012theano}. We used filter windows of 3, 4, and 5 with 100 feature maps each for the CNN. A dropout rate of $0.5$ and a batch size of 50 was employed, 10\% of the trainset was used as validation set and early stopping was adopted. Training is done through stochastic gradient descent over shuffled mini-batches with Adadelta update rule (we used an adadelta decay parameter of $0.95$). To initialise the models, GloVE word vectors were used~\cite{pennington2014glove} with a dimension $d=100$. System-action word-embeddings are tuned during training, instead hypothesis word-embeddings are not because of the heavy computations.

\subsection{Experiments}
\paragraph{Step I: Joint classification of dialogue-acts and slots:}
We evaluated five different model configurations for the joint classification of dialogue-acts and presence or absence of slots. 
\begin{itemize}
\item{\textbf{CNN}}: the softmax layers for the joint classification of dialogue acts and slots are connected directly to the CNN sentence representation with no context.  
\item{\textbf{CNN+LSTM}}: we study the influence of context by considering the previous system actions (Section~\ref{ss:ctxtrepr}, Eq.~\ref{eq:comb}), here we study the different context length, by using a context window of 1, 4, and all the previous system actions, namely \textbf{CNN+LSTM\_w1},  \textbf{CNN+LSTM\_w4} and  \textbf{CNN+LSTM\_w} respectively.
\item{\textbf{LSTM\_all}}: Finally, we study the impact of long distance dependencies, by using mainly the LSTM model, with the previous system actions as input, but we inject the sentence representation as the last LSTM input.
\end{itemize}
 
\paragraph{Step II: Classification of slot value pairs:}
We select the best model in step I for predicting the presence of slots, then for each slot present we predict the  value, by using again the best architecture from the previous step.

\subsection{Evaluation Metrics}
We evaluate the performance of our models by using the conventional metrics for classification, namely accuracy, precision, recall and F-measure (F1-score).  
%Accuracy  is the number of correct classifications divided by the number of instances in the evaluation set . Precision is the proportion of correct results among all the results returned by the model, while recall is the proportion of correct results among the gold semantics. F-measure (F1-score) is the harmonic mean of the precision and recall. 

In addition, we used the ICE score (Eq.~\ref{eq:ice}) between the hypotheses and the reference semantics (ie. ground-truth) to measure the overall quality of the distribution returned by the models\cite{thomson2008evaluating}. Let $U$ be the number of utterances and $W$ be the number of available semantic items. Given $u=1..U$ and $w=1...W$, let:
\begin{equation}
\begin{array}{l}
c_{uw} = \bigg \{
\begin{array}{l} 
 p\mbox{, the confidence assigned to the hypothesis that the } w^{th} \mbox{ semantic item is part of utterance } u ,\\ 
 0 \mbox{,  if none was assigned.}
\end{array}\\\\
%\]
%\[
\delta_{uw} = \bigg \{
\begin{array}{l}
1\mbox{,  if the } w^{th} \mbox{ item is in the reference semantics for } u, \\
 0\mbox{, otherwise}
\end{array}\\\\
%\]
\mbox{ and } N=\sum_{uw}\delta_{uw} \mbox{, be the total number of semantic items in the reference semantics.}\\\\
%\[
\mbox{ICE}= \frac{1}{N_w}\sum -\log (\delta_{uw} c_{uw} + (1-\delta_{uw})(1-c_{uw})
\label{eq:ice}
%\]
\end{array}
\end{equation}
\section{Results and Discussion}
\label{s:eval}
In this section we report the results on DSTC2 and In-car dialogue corpora.

\paragraph{Step I: Joint classification of dialogue-acts and slots:}
For this step, the classifiers must predict jointly 14 dialogue acts and 5 slots for the DSTC2 dataset as well as 14 dialogue acts and 7 slots for the In-car dataset.
We evaluate both (i) using 10 fold cross-validation on the trainsets and (ii) on the corpora' testsets.

Table~\ref{t:cvsIdstc2} shows the 10 fold cross-validation results on both corpora.
These results suggest that for DTSC2, the context representation is not significantly impacting the prediction. Although, the model with a window of 4 ,\textbf{CNN+LSTM\_w4}, improves slightly the accuracy and f1-score. On the In-car dataset, however,  including the context does help to disambiguate the semantic predictions from ill-formed hypotheses. This is expected, since this data set has a much higher error rate and hence higher levels of confusion in the ASR output. Although there is no significant difference on the f1-score when using the immediate previous system act ($w1$) or a longer context, % using all the context 
\textbf{CNN+LSTM\_w} gives a better accuracy and a lower ICE score on this dataset.

\begin{table}[htbp]
\centering
  \begin{tabular}{l | ll | lll  }
 \hline
 \scriptsize{Corpus}&\scriptsize{Metric}&\scriptsize{CNN}&\multicolumn{3}{|c}{\scriptsize{CNN+LSTM}}\\
 \hline
 \scriptsize{-}&\scriptsize{-}&\scriptsize{-}&\scriptsize{w1.}&\scriptsize{w4}&\scriptsize{w}\\\hline
 \scriptsize{DSTC2}&\scriptsize{acc.}&\scriptsize{$96.1\%\pm0.002$}&\scriptsize{$95.97\%\pm0.003$}&\scriptsize{$\mathbf{96.11}\%\pm0.002$}&\scriptsize{$95.9\%\pm0.003$}\\
& \scriptsize{P.}&\scriptsize{$90.17\%\pm0.007$}&\scriptsize{$89.33\%\pm0.007$}&\scriptsize{$89.77\%\pm0.004$}&\scriptsize{$89.21\%\pm0.008$}\\
 &\scriptsize{R.}&\scriptsize{$85.61\%\pm0.009$}&\scriptsize{$85.66\%\pm0.007$}&\scriptsize{$86.40\%\pm0.006$}&\scriptsize{$85.96\%\pm0.006$}\\
& \scriptsize{F1}&\scriptsize{$87.8\%\pm0.007$}&\scriptsize{$87.43\%\pm0.006$}&\scriptsize{$\mathbf{88.03}\%\pm0.004$}&\scriptsize{$87.53\%0.005$}\\
 &\scriptsize{ICE}&\scriptsize{$\mathbf{0.245\pm0.013}$}&\scriptsize{$0.275\pm0.02$}&\scriptsize{$0.271\pm0.02$}&\scriptsize{$0.277\pm0.02$}\\
 \hline
 \scriptsize{In-car}&\scriptsize{acc.}&\scriptsize{$90.45\%\pm0.005$}&\scriptsize{$91.66\%\pm0.003$}&\scriptsize{$91.49\%\pm0.007$}&\scriptsize{$\mathbf{91.77}\%\pm0.04$}\\
 &\scriptsize{P.}&\scriptsize{$83.87\%\pm0.01$}&\scriptsize{$84.31\%\pm0.01$}&\scriptsize{$84.16\%\pm0.01$}&\scriptsize{$83.89\%\pm0.01$}\\
 &\scriptsize{R.}&\scriptsize{$71.57\%\pm0.007$}&\scriptsize{$74.91\%\pm0.005$}&\scriptsize{$74.6\%\pm0.02$}&\scriptsize{$74.76\%\pm0.01$}\\
 &\scriptsize{F1}&\scriptsize{$76.96\%\pm0.008$}&\scriptsize{$\mathbf{79,16\%\pm0.003}$}&\scriptsize{$78.85\%\pm0.01$}&\scriptsize{$78.83\%\pm0.007$}\\
 &\scriptsize{ICE}&\scriptsize{$0.498\pm0.0013$}&\scriptsize{$0.457\pm0.02$}&\scriptsize{$0.459\pm0.03$}&\scriptsize{$\mathbf{0.448\pm0.02}$}\\
 \hline
 \end{tabular}
 \caption{10 fold cross-validation evaluation of step I, the joint classification of dialogue acts and slots. Here we study the impact of the context by comparing \textbf{CNN} and \textbf{CNN+LSTM}.
 %The star denotes statistical significance with the Wilcoxon test ($p < 0.005$)
 }\label{t:cvsIdstc2}
 \end{table}

Table \ref{t:tstset} shows the results on the test sets. Consequently, when evaluating on the DSTC2 test set, a window of 4 ($w4$), performs slightly better than other window sizes and better than the simple \textbf{CNN} model.
On the In-car testset, a context window of 4 outperforms all the other settings: \textbf{CNN+LSTM}. However, on this test set using the sentence
representation as the last input to the LSTM context neural network (section \ref{ss:comb})  improves the f1-score and reduces the ICE error. 
 
\begin{table}[htbp]
\centering
  \begin{tabular}{l | ll | lll | l}
 \hline
 \scriptsize{Corpus}&\scriptsize{Metric}&\scriptsize{CNN}&\multicolumn{3}{|c|}{\scriptsize{CNN+LSTM}}&\scriptsize{LSTM\_all}\\
 \hline
 \scriptsize{-}&\scriptsize{-}&\scriptsize{-}&\scriptsize{w1.}&\scriptsize{w4}&\scriptsize{w}&\scriptsize{-}\\\hline
 \scriptsize{DSTC2}&\scriptsize{acc.}&\scriptsize{$96.03\%$}&\scriptsize{$95.79\%$}&\scriptsize{$95.79\%$}&\scriptsize{$95.69\%$}&\scriptsize{$95.59\%$}\\
& \scriptsize{P.}&\scriptsize{$89.73\%$}&\scriptsize{$88.69\%$}&\scriptsize{$88.95\%$}&\scriptsize{$88.38\%$}&\scriptsize{$88.15\%$}\\
 &\scriptsize{R.}&\scriptsize{$84.74\%$}&\scriptsize{$85.09\%$}&\scriptsize{$86.02\%$}&\scriptsize{$85.96\%$}&\scriptsize{$84.76\%$}\\
& \scriptsize{F1}&\scriptsize{$87.14\%$}&\scriptsize{$86.83\%$}&\scriptsize{$\mathbf{87.43}\%$}&\scriptsize{$87.12\%$}&\scriptsize{$86.42\%$}\\
 &\scriptsize{ICE}&\scriptsize{$\mathbf{0.268}$}&\scriptsize{$0.278$}&\scriptsize{$0.292$}&\scriptsize{$0.297$}&\scriptsize{$0.308$}\\
 \hline
 \scriptsize{In-car}&\scriptsize{acc.}&\scriptsize{$\mathbf{87.60}\%$}&\scriptsize{$82.19\%$}&\scriptsize{$82.25\%$}&\scriptsize{$82.14\%$}&\scriptsize{$82.3\%$}\\
 &\scriptsize{P.}&\scriptsize{$69.96\%$}&\scriptsize{$79.52\%$}&\scriptsize{$79.29\%$}&\scriptsize{$80.25\%$}&\scriptsize{$78.12\%$}\\
 &\scriptsize{R.}&\scriptsize{$62.14\%$}&\scriptsize{$71.09\%$}&\scriptsize{$71.59\%$}&\scriptsize{$70.9\%$}&\scriptsize{$74.04\%$}\\
 &\scriptsize{F1}&\scriptsize{$65.53\%$}&\scriptsize{$74.89\%$}&\scriptsize{$75.15\%$}&\scriptsize{$75.02\%$}&\scriptsize{$\mathbf{75.9}\%$}\\
 &\scriptsize{ICE}&\scriptsize{$1.332$}&\scriptsize{$1.344$}&\scriptsize{$1.333$}&\scriptsize{$1.421$}&\scriptsize{$\mathbf{1.106}$}\\
 \hline
 \end{tabular}
 \caption{Evaluation of the Step I on DSTC2 and In-car testsets. We also compare two ways of combining sentence and context representation: \textbf{CNN+LSTM} models (combining sentence and context representation through a non linear function) and \textbf{LSTM\_all} model (embedding the sentence representation into the context model).}\label{t:tstset}
 \end{table}

\paragraph{Step II: Prediction of slot value pairs}
For evaluating Step II, we selected the best model obtained during the 10-fold cross-validation experiments in terms of F1 score. 
For both corpora, this was the \textbf{CNN+LSTM\_w4} configuration. 
For DSTC2, it was the $4^{th}$-fold crossvalidation with $Acc=90.42\%$, $F1=88.69\%$ and $\mbox{ICE}=0.251$. For In-car, it was the $5^{th}$-fold crossvalidation with $Acc=93.13\%$, $F1=81.49\%$ and $\mbox{ICE}=0.393$. We used these models to classify whether a given slot appears in a given hypothesis or not. Then for that slot, we train another \textbf{CNN+LSTM\_w4} classifier for predicting its values. In the In-car corpus the slot "type" has only one possible value "restaurant".  Similarly,  the slot "task" can only be the value "find". For these slots with only one value, we report values using the model of Step I, since it is enough to detect the slot in the utterance. %~\footnote{}. 

\begin{table}[ht]
\centering
  \begin{tabular}{l|lllll|lllll}
 \hline
 \scriptsize{}&\multicolumn{5}{|c|}{\scriptsize{DSTC2}}&\multicolumn{5}{|c}{\scriptsize{In-car}}\\\hline
 \scriptsize{Slot}&\scriptsize{Acc.}&\scriptsize{P.}&\scriptsize{R.}&\scriptsize{F1}&\scriptsize{ICE}&\scriptsize{Acc.}&\scriptsize{P.}&\scriptsize{R.}&\scriptsize{F1}&\scriptsize{ICE}\\
 \hline
 \scriptsize{Slot\footnote{"Slot" is used when no value is given for the slot (e.g., "What kind of food do they serve?"/request(slot=food)).}}&\scriptsize{95.29\%}&\scriptsize{90.89\%}&\scriptsize{95.72\%}&\scriptsize{93.24\%}&\scriptsize{0.478}&\scriptsize{89.92\%}& \scriptsize{74.73\%}& \scriptsize{61.56\%}& \scriptsize{67.51\%}& \scriptsize{ 0.743}\\
\scriptsize{Area}&\scriptsize{91.77\%}&\scriptsize{92.66\%}&\scriptsize{92.83\%}&\scriptsize{92.74\%}&\scriptsize{ 0.563}&\scriptsize{72.03\%}& \scriptsize{72.56\%}& \scriptsize{74.28\%}& \scriptsize{73.41\%}& \scriptsize{1.676}\\
\scriptsize{Food}&\scriptsize{71.37\%}&\scriptsize{73.19\%}&\scriptsize{76.02\%}&\scriptsize{74.58\%}&\scriptsize{1.989}&  \scriptsize{66.46\%}& \scriptsize{64.27\%}& \scriptsize{68.70\%}& \scriptsize{66.41\%}& \scriptsize{2.309}\\
\scriptsize{Price}&\scriptsize{94.62\%}&\scriptsize{91.33\%}&\scriptsize{94.49\%}&\scriptsize{92.89\%}&\scriptsize{ 0.729}&  \scriptsize{93.96\%}& \scriptsize{88.77\%}& \scriptsize{92.03\%}& \scriptsize{90.37\%}& \scriptsize{0.632}\\
\scriptsize{This\footnote{"This" is used for annotating elliptical utterances (e.g., "I dont care"/inform(this='dontcare')).}}&\scriptsize{98.70\%}&\scriptsize{96.79\%}&\scriptsize{93.92\%}&\scriptsize{95.33\%}&\scriptsize{0.113}&  \scriptsize{97.16\%}& \scriptsize{96.14\%}& \scriptsize{84.72\%}& \scriptsize{90.07\%}& \scriptsize{0.214}\\
\scriptsize{Type}&  \scriptsize{-}& \scriptsize{-} &\scriptsize{-} &\scriptsize{-}& \scriptsize{-}&\scriptsize{95.56\%}& \scriptsize{95.09\%} &\scriptsize{86.69\%} &\scriptsize{90.69\%}& \scriptsize{0.290}\\
 \scriptsize{Task}& \scriptsize{-}& \scriptsize{-}& \scriptsize{-}& \scriptsize{-}& \scriptsize{-}& \scriptsize{97.12\%}& \scriptsize{83.24\%}& \scriptsize{64.93\%}& \scriptsize{72.95\%}& \scriptsize{0.175}\\
 \hline
\scriptsize{Mean}&\scriptsize{90.35\%}&\scriptsize{88.97\%}&\scriptsize{90.60\%}&\scriptsize{89.76\%}&\scriptsize{0.774}&\scriptsize{87.47\%}& \scriptsize{82.11\%}& \scriptsize{76.13\%}& \scriptsize{78.77\%}& \scriptsize{0.863}\\
\scriptsize{St.Dev.}&\scriptsize{0.109}&\scriptsize{0.091}&\scriptsize{0.082}&\scriptsize{0.085}&\scriptsize{0.715}& \scriptsize{ 0.128}& \scriptsize{ 0.121}& \scriptsize{0.118}& \scriptsize{0.112}&  \scriptsize{ 0.821} \\\hline\hline
 \end{tabular}
 \caption{Evaluation of the step II: the slot-value pairs classification on DSTC2 and In-car.}
 \label{t:s2dstc2}
 \end{table}

% \begin{table}[htbp]
% \centering
%   \begin{tabular}{llllll}
%  \hline
%  \scriptsize{Slot}&\scriptsize{Acc.}&\scriptsize{P.}&\scriptsize{R.}&\scriptsize{F1}&\scriptsize{ICE}\\
%  \hline
%  \scriptsize{Slot}&\scriptsize{89.92\%}& \scriptsize{74.73\%}& \scriptsize{61.56\%}& \scriptsize{67.51\%}& \scriptsize{ 0.743}\\
%  \scriptsize{Area}&  \scriptsize{72.03\%}& \scriptsize{72.56\%}& \scriptsize{74.28\%}& \scriptsize{73.41\%}& \scriptsize{1.676}\\
%  \scriptsize{Food}&  \scriptsize{66.46\%}& \scriptsize{64.27\%}& \scriptsize{68.70\%}& \scriptsize{66.41\%}& \scriptsize{2.309}\\
%  \scriptsize{Price}&  \scriptsize{93.96\%}& \scriptsize{88.77\%}& \scriptsize{92.03\%}& \scriptsize{90.37\%}& \scriptsize{0.632}\\
%  \scriptsize{This}&  \scriptsize{97.16\%}& \scriptsize{96.14\%}& \scriptsize{84.72\%}& \scriptsize{90.07\%}& \scriptsize{0.214}\\
%  \scriptsize{Type}&  \scriptsize{95.56\%}& \scriptsize{95.09\%} &\scriptsize{86.69\%} &\scriptsize{90.69\%}& \scriptsize{0.290}\\
%  \scriptsize{Task}& \scriptsize{97.12\%}& \scriptsize{83.24\%}& \scriptsize{64.93\%}& \scriptsize{72.95\%}& \scriptsize{0.175}\\\hline
%  \scriptsize{Mean}&\scriptsize{87.47\%}& \scriptsize{82.11\%}& \scriptsize{76.13\%}& \scriptsize{78.77\%}& \scriptsize{0.863}\\
%  \scriptsize{St. Dev.} & \scriptsize{ 0.128}& \scriptsize{ 0.121}& \scriptsize{0.118}& \scriptsize{0.112}&  \scriptsize{ 0.821} 
%  \\\hline\hline
%  \end{tabular}
%  \caption{Evaluation of the step II: the slot-value pairs classification on In-car}
%  \label{t:s2incar}
%  \end{table}
 
Given that there is no domain specific delexicalisation, the models achieve a good level of performance overall (Table~\ref{t:s2dstc2}).  
Note that the slot "food"  has 74 possible values in DSTC2 and 25 in In-car.  Hence, this slot has much higher cardinality than all the other slots.

%respectively, reach acceptable  results, specially, taking into account that there was not delexicalisation.
 
 \paragraph{Overall performance}
 
A baseline for assessing overall performance is provided by the model presented in~\cite{Henderson2012a}, in which the vector representation is obtained by summing up the frequency of n-grams extracted from the 10-best hypotheses, weighted by their confidence scores. %(Eq.~\ref{eq:hmodel}). 
%  \begin{equation}
%  x_i=\sum_{j=1} C_{\mbox{hyp}_j} (\mbox{n-gram}_i).p_j
%  \label{eq:hmodel}
%  \end{equation}
Here we compare our performance against Henderson's model with and without context features, namely WNGRAMS+Ctxt and WNGRAMS repectively.
Henderson reported his results on the In-car dataset. A similar model, namely SLU1, was evaluated on DSTC2 in~\cite{williams2014web}. Both  implementations consist of many binary classifiers for dialogue act and slot-value pairs. 

\begin{table}[htbp]
\centering
  \begin{tabular}{llll}
 \hline
 \scriptsize{Corpus}&\scriptsize{Model}&\scriptsize{F1}&\scriptsize{ICE}\\\hline
 \scriptsize{DSTC2}&\scriptsize{SLU1~\cite{williams2014web}}&\scriptsize{$80.2\%$}&\scriptsize{$1.943$}\\
 &\scriptsize{CNN+LSTM\_w4}&\scriptsize{$\mathbf{83.59}\%$}&\scriptsize{$\mathbf{0.758}$}\\
 \scriptsize{In-car}&\scriptsize{WNGRAMS~\cite{Henderson2012a}}&\scriptsize{$70.8\%$}&\scriptsize{$1.76$}\\
&\scriptsize{WNGRAMS+Ctxt~\cite{Henderson2012a}}&\scriptsize{$\mathbf{74.2}\%$}&\scriptsize{$1.497$}\\
 &\scriptsize{CNN+LSTM\_w4}&\scriptsize{$73.06\%$}&\scriptsize{$\mathbf{1.106}$}\\\hline 
\end{tabular}
\caption{Overall performance of the setting CNN+LST\_w4 semantic decoder.}\label{t:overall}
 \end{table}
 In terms of the ICE score, the model CNN+LSTM W4 outperforms all the baselines (Table~\ref{t:overall}). In terms of the F1 score, the model significantly outperforms the SLU1 and WNGRAMS baselines. However it is slightly worse than WNGRAMS+Ctxt, which has been enhanced with context features on In-car. Remember however, that our model uses only word-embeddings for  automatically generating sentence and context representations without having any manually designed features or using explicit application specific semantic dictionaries.

\section{Conclusion and Future Work}
\label{s:disc}
This paper has presented a deep learning architecture for semantic decoding in spoken dialogue systems that exploits semantically rich distributed word vectors. We compared different models for combining sentence and context representations. We found that context representations significantly impact slot F-measure on ASR hypotheses generated under very noisy conditions. 
The combination of sentence and context representations, with a context window of 4 words, outperforms all the baselines in terms of the ICE score. In terms of the F1 scores, our model outperforms the baseline on the DSTC2 corpus and the baseline without manually designed features on the In-car corpus. Although the F-score of our model does not outperforms the baseline enriched with context features on the In-car corpus, the proposed model remains competitive, especially considering that our model requires no manually designed features or application specific semantic dictionaries.
%In the future, we want to study the adoption of sentence and context representations for \textit{Topic Management} in multi-domain dialogue systems. %(i.e., early detection of domain and users' intentions) 

\section{Future Work}
\label{s:fwork}
%In the future, we want to study the adoption of sentence and context representations for \textit{Topic Management} in multi-domain dialogue systems. 
Semantic distributed vector representations can be used for detecting similarity between domains. As future work, we want to study the adoption of the sentence and the contex representations generated in the Step I (i.e., the joint prediction of dialogue act and slots) within a \textit{Topic Management} in multi-domain dialogue systems. The Topic Manager is in charge of detecting the domain and the intention behind users' utterances. Furthermore, it would be interesting to study these embeddings for domain adaptation on potentially open-domains.

\section*{Acknowledgments}
This research was partly funded by the EP-SRC grant EP/M018946/1 Open Domain Statistical Spoken Dialogue Systems. The data used in this paper was produced in an industry funded project and it is not available for public use.

\bibliographystyle{acl}
\bibliography{coling2016}

\end{document}